\documentclass[10pt,twocolumn,letterpaper]{article}

\usepackage{cvpr}              
\usepackage{multirow}
\usepackage{tabularx}
\usepackage{lipsum} 
\usepackage{graphicx} 
\usepackage{adjustbox}
\usepackage{subcaption}
\usepackage{caption}  
\usepackage{tcolorbox}
\tcbuselibrary{skins}
\definecolor{cvprblue}{rgb}{0.21,0.49,0.74}
\usepackage[pagebackref,breaklinks,colorlinks,allcolors=cvprblue]{hyperref}

\def\paperID{****}
\def\confName{CVPR}
\def\confYear{2026}
\title{DialogueVPR: Towards Conversational Visual Place Recognition}

\author{
    Yukun Song\textsuperscript{1,$*$} \quad 
    Changwei Wang\textsuperscript{2, $*$} \quad 
    Xingtian Pei\textsuperscript{1,$*$} \quad 
    Shibiao Xu\textsuperscript{1}$\dagger$ \\
    Wenhao Xu\textsuperscript{1} \quad 
    Shunpeng Chen\textsuperscript{1} \quad 
    Yu Zhang\textsuperscript{3} \quad 
    Ke Zhang\textsuperscript{1} \\
    Rongtao Xu\textsuperscript{4} \quad 
    Xuxiang Feng\textsuperscript{5,6}$\dagger$ \quad 
    Pengyang Wang\textsuperscript{5} \\
    \\
    \textsuperscript{1}School of Artificial Intelligence, Beijing University of Posts and Telecommunications \\
    \textsuperscript{2} Key Laboratory of Computing Power Network and Information Security, Ministry of Education, \\ \quad Shandong Computer Science Center, Qilu University of Technology \quad  \textsuperscript{3} Macquarie University \quad \\ \textsuperscript{4} Spatialtemporal AI \quad
    \textsuperscript{5} University of Macau  \textsuperscript{6} Aerospace Information Research Institute \\
    {\tt\small \{shibiaoxu@bupt.edu.cn, fengxx@aircas.ac.cn\}}
}
\begin{document}
\twocolumn[{%
\renewcommand\twocolumn[1][]{#1}%
\maketitle  

\begin{center}
    \centering
    \vspace{-0.4cm} 
    \includegraphics[width=\textwidth]{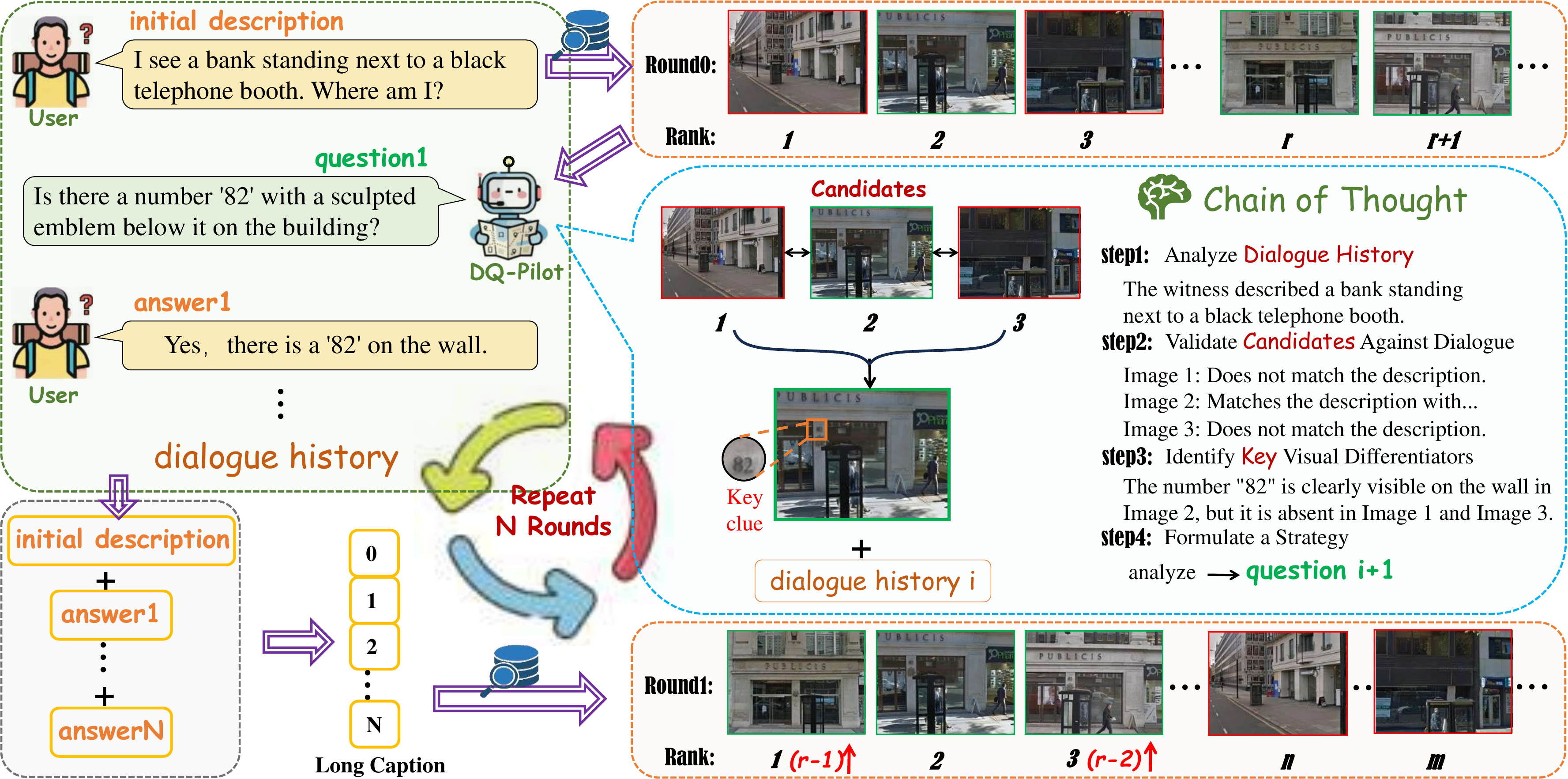}
    \captionof{figure}{
    \textbf{An Illustration of the Dialogue Place Recognition (DlgPR) Framework.}
    The key component, DQ-Pilot, functions as a reasoning agent that transforms geolocalization from a simple one-shot "retrieval" into a sophisticated "reasoning-based retrieval" process: the user provides an "initial description", and the retriever CMPL performs a preliminary retrieval (Round 0) to generate multiple visually similar candidate locations. DQ-Pilot formulates high-information-gain questions regarding the "candidate locations" and "dialogue history" through a chain of thought. The user responds to this question with crucial new information, and such feedback is integrated into the dialogue history to form a more detailed context for the subsequent retrieval round. This iterative loop of "analysis-questioning-optimization" enables the system to progressively resolve ambiguities and accurately identify the target location.
    }
    \label{fig:main_figure}
    \vspace{0.3cm} 
\end{center}%
}]
{\let\thefootnote\relax\footnotetext{$^\dagger$Corresponding authors. $^*$Equal contribution. }}
\begin{abstract}
Inspired by how humans communicate spatial information, language-guided geo-localization has gained significant traction for its intuitive and practical value. Despite this progress, most methods still rely on a static, one-shot retrieval paradigm, which fails to handle the ambiguity and incompleteness inherent in real-world natural language descriptions. We propose a paradigm shift to reasoning retrieval and introduce Dialogue Place Recognition (DlgPR), which casts localization as an interactive, dialogue-driven reasoning process. To support this new task, we present DlgQuest-Cities, the first large-scale dialogue-based benchmark for place recognition, and a unified reasoning framework that couples a cross-modal multi-level retriever with an intelligent questioner, DQ-pilot. DQ-pilot is trained in a curriculum: supervised fine-tuning on a curated DQ-cities-20k subset  followed by reinforcement refinement on a harder DQ-cities-10k split via GRPO. Two task-aligned metrics guide learning: a Discriminative Difficulty Index (DDI) for curriculum sampling and a Positional Retrieval Gain (PRG) reward that directly measures retrieval improvement induced by a question. Experiments show this reasoning-based approach significantly outperforms baselines. 
The code and model are available at https://github.com/Graysonggg/DlgPR.
\end{abstract}    
\section{Introduction}
\label{sec:intro}

Accurately perceiving and determining one’s location remains a fundamental challenge for both humans~\cite{tian2024loc4plan} and intelligent agents~\cite{sarlin2019coarse}. Solving this problem underpins a wide range of applications, including precise pedestrian navigation in urban environments, autonomous robot operation in dynamic scenes, and localization correction in GPS-denied areas such as urban canyons~\cite{wang2023fine}. Motivated by these demands, community's recent research has explored a more intuitive paradigm—place recognition driven by natural language descriptions~\cite{wang2023fine,kolmet2022text2pos,xia2024text2loc,ye2024skydiffusion,hu2024progeo}. Reflecting everyday human interactions, these approaches holds strong practical value: a passenger verbally guiding a taxi driver~\cite{ye2025cross}, identifying a place through spoken directions, or describing the surroundings in an emergency call~\cite{chen2024scene}, or commanding a home service robot through natural language~\cite{zhang2024navid}.


Recent language-driven localization methods, such as Text2Pose~\cite{kolmet2022text2pos} and Text2Loc~\cite{xia2024text2loc}, primarily focus on identifying individual locations within 3D point clouds. However, constructing and storing large-scale 3D maps remains costly, hindering practical deployment. Instead, recent work~\cite{ye2025cross,lyu2024tell} frames the problem as a large-scale retrieval task by correlating natural language with expansive, readily available visual data like satellite or street-view images. Despite progress, most language-guided localization methods still follow a static retrieval paradigm, where a fixed textual query is processed once to return the best-matching location. The fundamental limitation of this design lies in its \textit{passivity}: it fails to handle the ambiguity inherent in real-world descriptions. When the initial input is vague or incomplete—such as an imprecise verbal account (the “user description dilemma”) or an erroneous recollection—these systems cannot actively seek clarification or gather additional information. Consequently, single-turn, non-interactive retrieval remains fragile in dynamic, real-world scenarios.

To transcend these constraints, we argue that geo-localization should evolve from passive retrieval to an advanced paradigm of reasoning Retrieval. An intelligent agent must move beyond passive matching toward active understanding, reasoning, and interaction with uncertain environments and ambiguous human instructions.



To drive this paradigm shift, we introduce Dialogue Place Recognition (DlgPR)—a new task that reformulates localization as an iterative, collaborative dialogue. In DlgPR, the system transforms from a passive retriever into an active reasoner: it analyzes candidate locations, proactively engages the user with targeted questions to obtain discriminative evidence, and incrementally refines its belief about the correct place as the dialogue history becomes richer and the information more complete. Specifically, we develop a unified reasoning framework composed of a Cross-Modal Progressive Learning (CMPL) Retriever and an intelligent Multimodal Large Language Model, Dialogue-Quest-Pilot (DQ-pilot). The CMPL retriever is responsible for iteratively integrating information from the evolving dialogue to refine its search and retrieve relevant candidate locations. These candidates are then passed to DQ-pilot, which acts as the reasoning core—diagnosing ambiguity and generating questions to maximize information gain. This synergy transforms the system from a passive retriever into an active reasoner, enabling efficient and precise localization by incrementally refining its belief as the dialogue unfolds.

Our main contributions are summarized as follows:
\begin{itemize}
\item We propose a novel task, Dialogue Place Recognition (DlgPR), which shifts the paradigm from static retrieval to active, dialogue-driven reasoning. To facilitate research on this new task, we construct DlgQuest-Cities (DQ-cities), the first large-scale benchmark dataset for dialogue-based place recognition.
\item We develop DlgQuest, a unified and effective reasoning framework featuring a cross-modal retriever (CMPL) and an MLLM agent (DQ-Pilot). Crucially, to train this framework, we introduce a novel curriculum learning strategy guided by two task-aligned metrics—a Discriminative Difficulty Index (DDI) and a Positional Retrieval Gain (PRG)—enabling the agent to learn progressively from basic perception to advanced reasoning. Extensive experiments demonstrate the superiority of our approach.
\end{itemize}

\section{Related Work}
\label{sec:formatting}

\subsection{Natural Language-Driven Visual Perception and Localization}

Geo-localization \citep{NetVLAD,SALAD,BoQ,SuperVLAD,EMVP,SelaVPR,FoL,shi2020looking,salad-cm,garg2022seqmatchnet,deuser2023sample4geo,ali2023global,li2023omnicity,xia2024adapting,ye2024cross,SAGE,chen2026region} predicts a query's location by retrieving similar images from a geo-tagged database. Recently, multi-modal retrieval incorporating natural language has emerged in this field \citep{hong2019textplace,lyu2024tell,hu2024progeo,wang2024lvlm,shang2025bridging,tao2025textinplace}. For example, \citep{ye2025cross} introduces scene text, breaks through the limitation of text length, and, for the first time, introduces an interpretability framework, ensuring the localization process is no longer a black box. Meanwhile, \citep{,chu2024towards} enhances the model's spatial perception capabilities by learning phrases that describe fine-grained spatial relationships in natural language through Blending Spatial Matching. In 3D localization, \citep{kolmet2022text2pos} uses natural language instructions for position matching in point clouds, while \citep{xia2024text2loc} advances this by directly fusing textual semantics with geometric features for end-to-end position regression. For indoor recognition, \citep{tao2025textinplace} refines ranking using discriminative text filtered from images. Despite their success, these text-driven geolocation tasks remain largely static and lack dynamic interaction capabilities.

\begin{figure*}[tb]
  \centering
  \includegraphics[width=\textwidth]{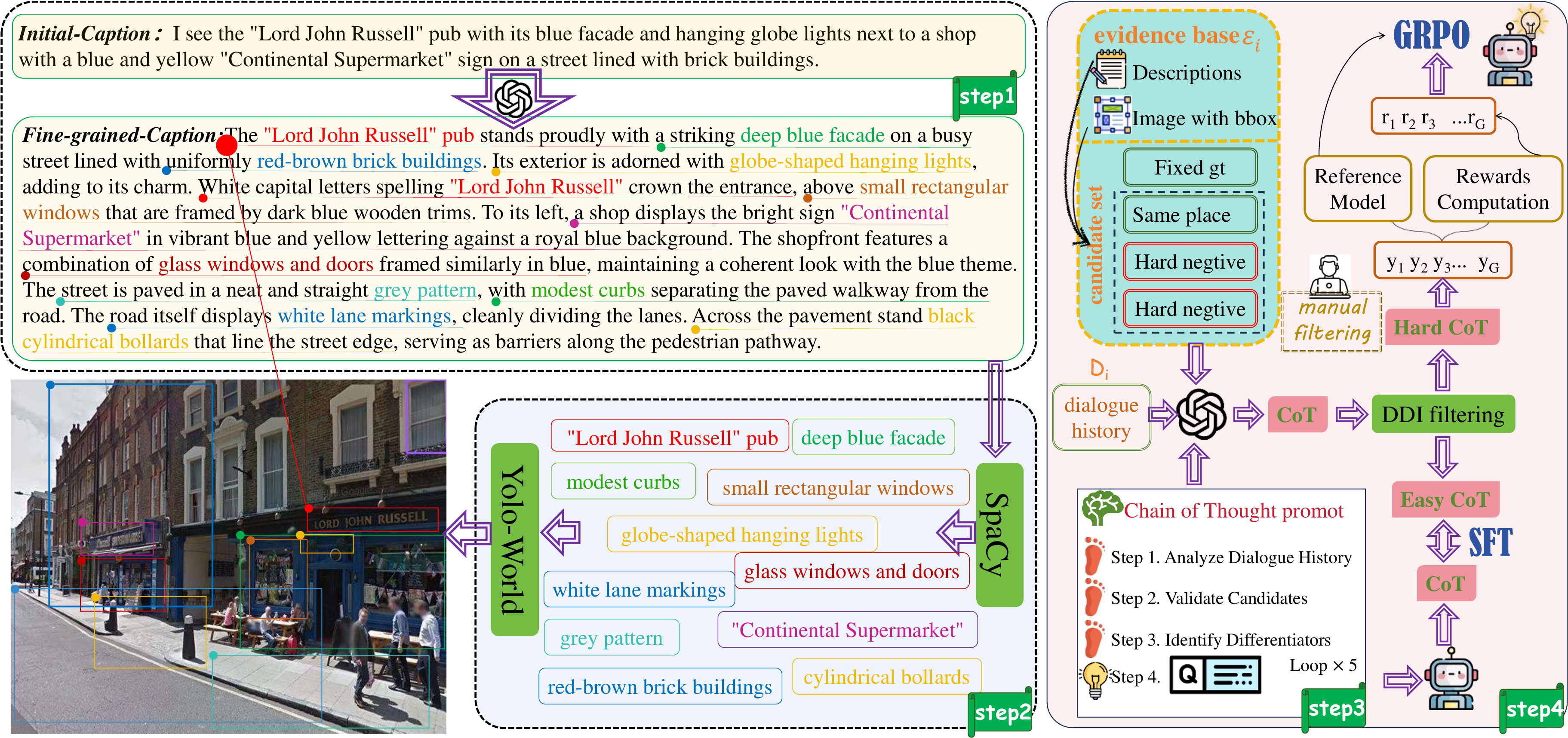} 
  \caption{Dataset Construction Flowchart. It is mainly divided into 4 parts: Text modality expansion; Region-level visual evidence construction; Chain-of-Thought dialogue generation; Discriminative difficulty-aware sampling. }
  \vspace{-1em}
  \label{fig:data-construction}
\end{figure*}


\subsection{Interactive Retrieval}
Cross-modal interactive retrieval has been actively explored in text-to-image \citep{levy2023chatting,lee2024interactive,zhu2024enhancing,lu2025llava} and text-to-video domains \citep{madasu2022learning, liang2023simple}, encompassing various interaction formats \citep{kovashka2015whittlesearch, cai2021ask,lee2021cosmo}. For example, \citep{liang2023simple} diversifies question generation, while PlugIR \citep{lee2024interactive} decouples dialogue understanding from retrieval via LLMs, enabling compatibility with black-box models. Furthermore, LLaVA-ReID \citep{lu2025llava} generates questions maximizing information gain through forward-looking supervision.

Ultimately, interactive retrieval aims to replicate human-like logical reasoning. However, current multi-turn dialogue methods primarily perform reactive information aggregation based on explicit feedback, lacking deeper proactive reasoning capabilities. In contrast, our work pioneers the first multi-modal interactive reasoning task in the field of geolocation.



\subsection{Visual Reinforcement Learning}

The advent of the OpenAI’s o1 \citep{jaech2024openai} and DeepSeek-R1 reasoning model \citep{guo2025deepseek} introduced the paradigm of incorporating visual reasoning into visual tasks. Reinforcement learning (RL) is pivotal for endowing models with reasoning capabilities, and Group Relative Policy Optimization (GRPO) \citep{shao2024deepseekmath}, characterized by its verifiable rewards, has emerged as a prominent RL methodology. Building on this, VLM-R1 \citep{shen2025vlm} developed multiple verifiable reward functions to fine-tune Vision-Language Models (VLMs). Subsequently, Visual-RFT \citep{liu2025visual} formulated simple yet effective reward functions for diverse visual tasks, further enabling efficient learning under data-scarce conditions.
Existing research demonstrates that, compared to Supervised Fine-Tuning (SFT), GRPO facilitates deeper reasoning, offers greater interpretability through its reasoning process, and exhibits superior generalization under limited supervision. Therefore, our proposed framework, DlgQuest, employs both SFT and GRPO to achieve active, reasoning-based geolocation.

\section{DlgQuest-Cities}


\subsection{Overview}


To support the dialogical reasoning required by our proposed DlgPR task, we construct the DlgQuest-Cities (DQ-cities) dataset. This new benchmark is built upon the widely-used GSV-Cities collection~\cite{gsv}, augmenting its rich geo-tagged imagery with multi-layered annotations tailored for dialogue-based localization. Each location in DlgQuest-Cities is annotated with information specifically designed for interactive spatial reasoning. Specifically, the dataset includes:
(1) Initial ambiguous place captions, simulating users’ vague or uncertain verbal queries based on incomplete memories; (2) Fine-grained place descriptions, offering comprehensive visual semantic details that serve as the factual foundation for multi-turn reasoning; (3) Region-level annotations, where bounding boxes are paired with corresponding textual descriptions to provide localized evidence for spatial grounding; and (4) Multi-turn, goal-oriented dialogues, in which each question is purposefully designed to differentiate visually similar locations and progressively resolve ambiguity. DQ-cities in total consists of 106,880 location images and 30k carefully selected conversation samples. Each fine-grained description has an average of 154.6 words, with the maximum reaching up to 262 words.

\subsection{Dataset Construction}
The rich annotations in DlgQuest-Cities are generated via an automated, multi-stage pipeline designed to produce the textual and dialogical data needed to train DQ-pilot. This pipeline, illustrated in Fig.\ref{fig:data-construction}, is specifically engineered to synthesize strategy-aware dialogues for each place. It consists of four principal stages: (1) text-modality expansion, (2) text-driven region-level annotation, (3) Chain-of-Thought (CoT) based dialogue generation guided by GPT-4o and (4) Curriculum sampling based on discrimination difficulty.

\textbf{Step 1: Text Modality Expansion.} This initial stage is responsible for creating the foundational textual layers for each place. To emulate a user's initial query, the pipeline first generates an initial ambiguous caption (e.g., “I see a bank with a telephone booth beside it.”). Following this, the system expands the caption into a long-form, fine-grained place description. This detailed narrative serves as a fact-rich foundation for subsequent dialogue generation. The generation process is constrained by a structured prompt (see Appendix) with task-specific rules: it focuses strictly on static elements (e.g., buildings, signage, spatial relationships), disregards transient objects (e.g., cars), and emphasizes features informative for place recognition. This procedure ensures the resulting descriptions provide reliable, factually grounded information for downstream reasoning.

\textbf{Step 2: Region-Level Visual Evidence Construction}
To ground the dialogue in specific visual details, DlgQuest-Cities incorporates region-level annotations. We begin by extracting salient noun phrases from the fine-grained descriptions. Unlike approaches such as FG-CLIP~\cite{xie2025fg} that often rely on simple nouns, we employ a greedy expansion strategy with \textit{spaCy} to capture maximally descriptive phrases, including rich adjectival modifiers and prepositional clauses (e.g., “the red brick bank with green awnings”). These descriptive phrases serve as more effective text prompts for an open-vocabulary detector (YOLO-World), enabling it to localize the corresponding objects with greater precision. The final output is a set of structured annotations, where each annotation links a specific image region (the bounding box) to its corresponding textual phrase. This step enriches the visual evidence base of the place, effectively avoiding the omission of key details by the teacher model.


\textbf{Step 3: Chain-of-Thought Dialogue Generation}
This stage constructs the interactive reasoning samples that power the DlgPR training process. For each place, we synthesize a five-round dialogue sequence, where each round simulates one reasoning–questioning cycle of the teacher model.

At each dialogue round \(i\), the pipeline assembles a decision-making context composed of: (1) a compact and distinctive candidate set—comprising the target image \(I_{\mathrm{t}}\), positive samples \(I_{\mathrm{p}}\) from the same location, and two challenging negatives \(I_{\mathrm{n1}}, I_{\mathrm{n2}}\) retrieved by trained CMPL; (2) the evidence base \(\mathcal{E}_i = \{(I_{\mathrm{t}}, t_{\mathrm{t}}, B_{\mathrm{t}}), (I_{\mathrm{p}}, t_{\mathrm{p}}, B_{\mathrm{p}}), (I_{\mathrm{n}}, t_{\mathrm{n}}, B_{\mathrm{n}})\}\), where \(t\) and \(B\) denote the textual descriptions and bounding boxes obtained in Steps 1 and 2; and (3) the accumulated dialogue history \(D_i\).


Next, the teacher model (GPT-4o) is prompted to execute a four-step chain-of-thought before composing the next question:
\begin{itemize}[leftmargin=1.5em]
    \item \textbf{Analyze Dialogue History:} summarize confirmed and ruled-out evidence contained in \(D_i\);
    \item \textbf{Validate Candidates Against Dialogue:} compare each candidate’s evidence in \(\mathcal{E}_i\) with \(D_i\) and eliminate inconsistent ones;
    \item \textbf{Identify Key Visual Differentiators:} examine the remaining candidates within \(\mathcal{E}_i\) to pinpoint region-grounded, text-anchored cues that most clearly distinguish them;
    \item \textbf{Formulate a Strategy:} Design a question that targets the most decisive visual uncertainty to maximize information gain.
\end{itemize}

Finally, the teacher’s internal deliberation and proposed question are wrapped in \(\langle think \rangle \langle /think \rangle\) and \(\langle question \rangle \langle /question \rangle\).

\textbf{Step 4: Discriminative Difficulty-Aware Curriculum Sampling}
\label{sec:ddi_sampling}
To ensure DQ-pilot learns progressively from simple to complex scenarios, we introduce a curriculum-aware sampling strategy. This strategy is guided by a unified Discriminative Difficulty Index (DDI), a weighted score combining two complementary metrics: Semantic Ambiguity (SA) and Retriever-Informed Difficulty (RID).

\textbf{Semantic Ambiguity (SA).}
SA quantifies the intrinsic ambiguity of a candidate set.  Given positive and negative textual embeddings \(t_{\mathrm{p}}\) and \(t_{\mathrm{n}}\), and their corresponding visual embeddings, we compute:
\begin{equation}
\begin{split}
\mathrm{SA} = {}& \alpha \cdot \mathrm{sim}\!\big(\phi_T(t_{\mathrm{t}}), \phi_T(t_{\mathrm{n}})\big) \\
                & + (1-\alpha) \cdot \big(1 - \mathrm{sim}(\phi_T(t_{\mathrm{t}}), \phi_T(t_{\mathrm{p}}))\big).
\end{split}
\end{equation}
where \(\phi_T(\cdot)\) is the text encoder of CMPL. A higher SA indicates stronger semantic overlap and thus greater ambiguity among candidates.

\textbf{Retriever-Informed Difficulty (RID).} RID measures the empirical difficulty of a dialogue turn by quantifying the rank improvement of positive samples after answering the generated question. Let $r_j^{(i-1)}$ and $r_j^{(i)}$ be the rank of a positive item $j \in \mathcal{P}$ before and after dialogue round $i$. The \textit{Positional Retrieval Gain} (PRG) normalizes the observed rank improvement against the maximum possible improvement:
\begin{equation}
\mathrm{PRG}_i = \frac{G^{(i)} - G^{(i-1)}}{G^{*} - G^{(i-1)}},
\end{equation}
where gain $G$ is the sum of nDCG-style~\cite{wang2013theoretical} contributions $c(r)=1/\log_2(r+1)$ over all items in $\mathcal{P}$, and $G^{*}$ represents the ideal total gain if all positive items occupied the top ranks ($G^{*}=\sum_{k=1}^{|\mathcal{P}|} c(k)$). We then set $\mathrm{RID}_i = 1 - \mathrm{PRG}_i$, so that minimal rank improvement (low PRG) corresponds to high empirical difficulty.

\textbf{DDI-based Curriculum Sampling.}
We first filter out low-quality dialogues (e.g., with minimal rank changes, $\mathrm{PRG}_i < \tau_1$) and overtly noisy ones using automated metrics. This automated screening is complemented by a brief manual inspection, primarily focused on borderline cases, to ensure overall data integrity. For the resulting filtered pool, we compute the final difficulty score:
\begin{equation}
\mathrm{DDI} = w_{\mathrm{sa}}\cdot\mathrm{SA} + w_{\mathrm{rid}}\cdot\mathrm{RID}.
\end{equation}
Using a threshold on the DDI score, we construct a two-stage curriculum:
\begin{itemize}[leftmargin=1.5em]
    \item \textbf{Stage 1 (For Supervised Fine-Tuning):} We sample 20k instances as DQ-cities-20k, prioritizing low-DDI samples ($\sim\!70\%$). This stage focuses on learning fundamental visual grounding and core reasoning patterns.
    \item \textbf{Stage 2 (For Reinforcement Learning):} We sample 10k instances as DQ-cities-10k, prioritizing high-DDI samples ($\sim\!70\%$). This stage challenges the model with highly ambiguous and hard-to-distinguish cases.
\end{itemize}

This entire pipeline, from description generation to curriculum sampling, produces the final 30k dialogue rounds in DQ-Cities. The resulting dataset is not only rich in content but also structured to facilitate progressive learning, advancing the model from basic visual grounding to robust, evidence-backed reasoning. More dataset statistics and construction details are provided in Appendix.

\section{Method}
\label{sec:formatting}

\begin{figure*}[tb]
  \centering
  \includegraphics[width=\textwidth]{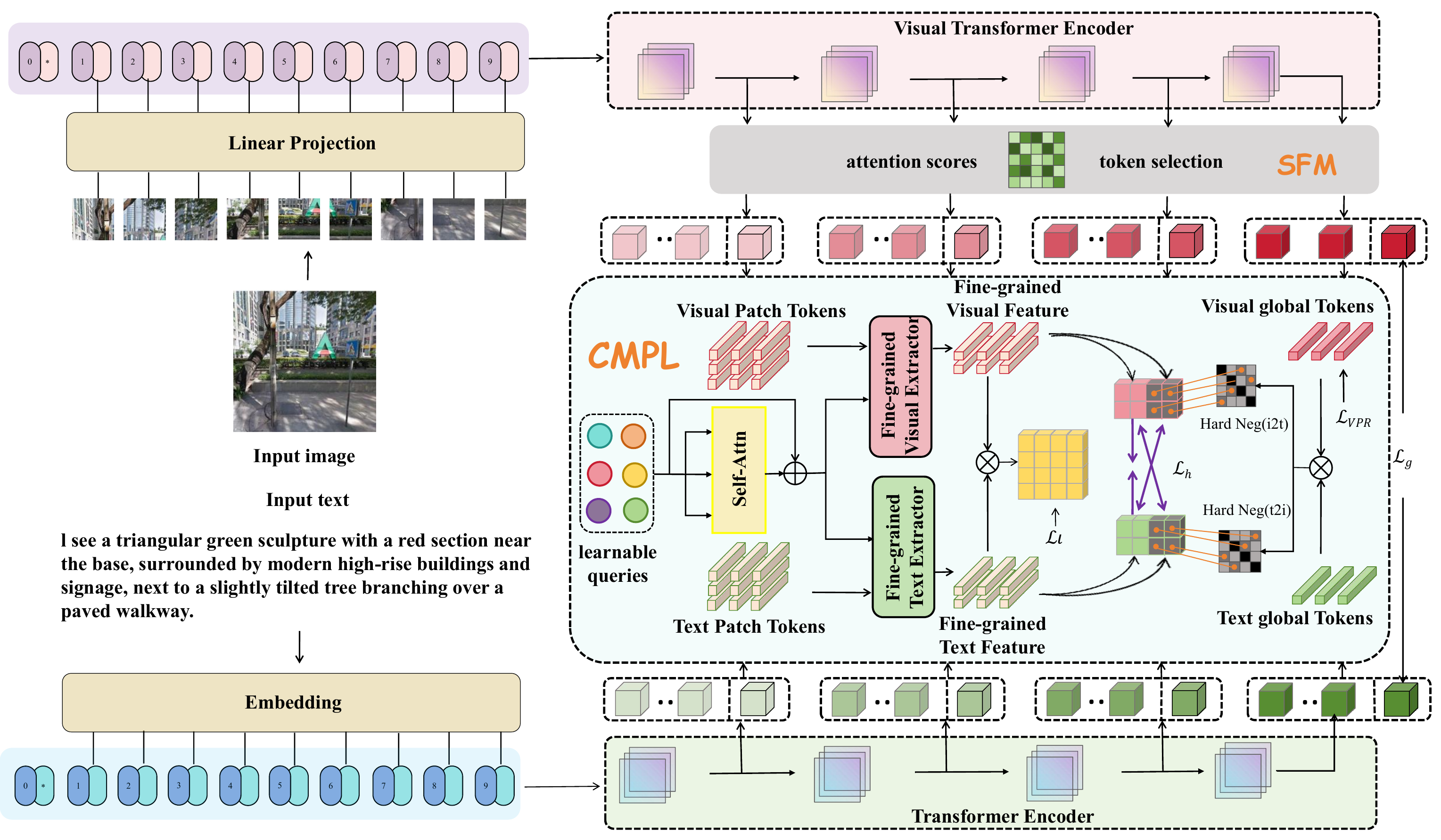} 
  \caption{Flowchart of the proposed cross-modal progressive learning retriever. The core is the cross-modal progressive learning (CMPL) module, which aligns the global and local information of multi-level visual and textual features respectively, and mines hard negative samples for triplet loss learning.}
  \vspace{-1em}
  \label{fig:Retriever}
\end{figure*}


\subsection{The Dialogue Place Recognition Framework}

The DlgPR framework reframes place recognition as a dynamic, interactive reasoning process, departing from traditional static retrieval. It orchestrates two core components: a multi-modal retriever, CMPL, that iteratively refines the search and a dialogue agent, DQ-pilot, that generates discriminative questions to resolve ambiguity.

The process begins when an initial user query, \(d_0\), yields a coarse set of candidate locations \(C_0\) via the CMPL retriever. To disambiguate these candidates, the system enters an iterative loop. At each round \(t\), DQ-pilot analyzes the current candidates \(C_t\) to formulate an optimal question \(q_t\). Upon receiving the user's answer \(a_t\), the framework aggregates the dialogue history into an enriched textual query \(d_{t+1} = \text{concat}(d_0, a_1, \dots, a_t)\) for the CMPL retriever. This updated query \(d_{t+1}\) enables CMPL to perform a more informed retrieval, producing a refined candidate set \(C_{t+1}\). This cycle of question-answering and retrieval progressively narrows the search space, achieving robust localization by resolving ambiguities through natural conversation.




\subsection{Cross-Modal Progressive Learning Retriever}
To support dialogue-driven reasoning, our Cross-Modal Progressive Learning (CMPL) retriever incrementally refines visual-textual alignment from local to global granularity. 

\noindent\textbf{Progressive Feature Alignment.} We extract hierarchical visual patches $V^{(l)}$ and text tokens $T^{(l)}$ from intermediate layers $P = \{p_3, p_6, p_9, p_{12}\}$. To highlight geographically relevant cues, $V^{(l)}$ is refined into $V_{s}^{(l)}$ via a saliency filtering module (SFM) that dynamically selects discriminative tokens based on attention weights, supervised by an auxiliary loss $L_{vpr}$~\cite{wang2019multi}.

To bridge modality structures, we introduce a shared fine-grained extractor $E_f$ and learnable \textit{instance-concept queries} $Q^{(l)}$. Acting as semantic anchors, they distill $V_{s}^{(l)}$ and $T^{(l)}$ into unified representations:
\begin{equation}
F_v^{(l)} = E_f(Q^{(l)}, V_{s}^{(l)}), \quad F_t^{(l)} = E_f(Q^{(l)}, T^{(l)}).
\end{equation}

\noindent\textbf{Hierarchical Similarity Distribution Matching.} We apply an SDM loss~\cite{jiang2023cross} at multiple granularities to minimize the bidirectional KL-divergence between the predicted similarity distribution $p$ and the ground-truth $q$. For an image anchor $F_{v,i}$, its predicted distribution across $B$ batch texts is:
\begin{equation}
p_{v \to t, i, j} = \frac{\exp(s_{i,j} / \tau)}{\sum_{k=1}^{B}\exp(s_{i,k} / \tau)},
\end{equation}
where $s_{i,j}$ is the similarity score and $\tau$ is the temperature. The target $q$ is normalized from binary batch labels.

\textbf{Hard-Negative Isolation (HI).}
To further improve geometric separability, we propose a Hard-Negative Isolation (HI) loss that applies localized repulsion to the most confusing negatives within each batch. For an image–text pair \((F_{v,i}, F_{t,i})\), the hardest negatives \(j^*\) and \(k^*\) are selected by similarity, and the margin-based triplet objective
\(L_{\text{hi}} = [d(F_{v,i},F_{t,i})^2 - d(F_{v,i},F_{t,j^*})^2 + \alpha]_+ + [d(F_{t,i},F_{v,i})^2 - d(F_{t,i},F_{v,k^*})^2 + \alpha]_+\)  
enforces discriminative separation across modalities.

\begin{table*}[t]
    \centering
    \caption{Interactive multi-round retrieval performance across five representative regions. We report the recall at the 3rd and 5th rounds (initiated from a short initial query), along with the BRI evaluation metric. The best metrics are shown in \textbf{\textcolor{red}{red bold}}.}
    \label{tab:city_results_modified}
    \setlength{\tabcolsep}{1.5mm}
    \renewcommand{\arraystretch}{1.1}
    
    \begin{tabularx}{\textwidth}{l l||*{2}{>{\centering\arraybackslash}X}||*{2}{>{\centering\arraybackslash}X}||*{2}{>{\centering\arraybackslash}X}||*{2}{>{\centering\arraybackslash}X}||*{2}{>{\centering\arraybackslash}X}||>{\centering\arraybackslash}X}
    
    \toprule
    
    \multirow{2}{*}{Method} & \multirow{2}{*}{Round}
      & \multicolumn{2}{c||}{\textbf{LosAngeles}}
      & \multicolumn{2}{c||}{\textbf{BuenosAires}}
      & \multicolumn{2}{c||}{\textbf{MexicoCity}}
      & \multicolumn{2}{c||}{\textbf{Osaka}}
      & \multicolumn{2}{c||}{\textbf{PRG}}
      & \multicolumn{1}{c}{\multirow{2}{*}{\textbf{BRI}${\downarrow}$}} \\
    
    \cline{3-12}
      & & \small{R@1} & \small{R@5}
      & \small{R@1} & \small{R@5}
      & \small{R@1} & \small{R@5}
      & \small{R@1} & \small{R@5}
      & \small{R@1} & \small{R@5}
      & \\ 
      
    \hline
    
    \multirow{1}{*}{Initial} 
    & round0 & 35.9 & 56.3 & 39.0 & 60.8 & 42.0 & 64.0 & 35.6 & 55.4 & 52.8 & 74.2 & / \\
    
    \hline
    \multirow{2}{*}{Qwen2.5-VL-7B} 
    & round3 & 42.4 & 59.3 & 45.5 & 64.1 & 47.2 & 65.9 & 41.2 & 57.9 & 58.0 & 74.6 &  \multirow{2}{*}{1.58} \\
    & round5 & 43.2 & 60.1 & 46.2 & 64.6 & 48.2 & 66.8 & 42.1 & 58.5 & 59.1 & 74.9 &  \\
    \hline
    \multirow{2}{*}{Qwen2.5-VL-72B} 
    & round3 & 46.1 & 65.2 & 49.3 & 69.2 & 52.4 & 71.7 & 46.2 & 64.7 & 62.9 & 80.0   & \multirow{2}{*}{1.44} \\
    & round5 & 49.5 & 68.6 & 51.9 & 71.4 & 54.6 & 74.0 & 49.1 & 67.5 & 65.1 & 82.1 &  \\
    \hline
    \multirow{2}{*}{PlugIR} 
    & round3 & 48.1 & 67.0 & 50.3 & 70.9 & 52.5 & 72.2 & 47.3 & 65.6 & 63.9 & 80.2 &  \multirow{2}{*}{1.41} \\
    & round5 & 51.2 & 70.3 & 53.2 & 72.5 & 55.7 & 75.1 & 50.5 & 68.8 & 66.2 & 83.2 & \\
    \hline
    \multirow{2}{*}{DlgQuest (SFT)} 
    & round3 & 49.2 & 68.4 & 51.8 & 71.5 & 53.6 & 73.7 & 48.1 & 66.9 & 64.1 & 81.5 & \multirow{2}{*}{1.29} \\
    & round5 & 54.6 & 73.6 & 55.9 & 74.7 & 57.8 & 77.4 & 53.3 & 71.6 & 68.4 & 85.3 &  \\
    \hline
    \multirow{2}{*}{DlgQuest (SFT+GRPO)} 
    & round3 & 52.1 & 71.6 & 54.0 & 75.5 & 58.0 & 76.9 & 52.9 & 71.9 & 67.8 & 84.4 & \multirow{2}{*}{\textbf{\textcolor{red}{1.18}}} \\
    & round5 & \textbf{\textcolor{red}{58.4}} & \textbf{\textcolor{red}{76.5}} & \textbf{\textcolor{red}{59.3}} & \textbf{\textcolor{red}{79.6}} & \textbf{\textcolor{red}{61.8}} & \textbf{\textcolor{red}{80.1}} & \textbf{\textcolor{red}{58.6}} & \textbf{\textcolor{red}{76.7}} & \textbf{\textcolor{red}{71.4}} & \textbf{\textcolor{red}{86.6}} & \\

    \bottomrule
    \end{tabularx}
    \vspace{-0.3cm}
\end{table*}

\textbf{Overall Objective.}
We apply this hierarchically. The global loss ($L_{gs}$) uses cosine similarity between [CLS] tokens. For local losses ($L_{ls}^{(l)}$) from a set of intermediate layers $P = \{p_3, p_6, p_9, p_{12}\}$, the score $s_{i,j}^{(l)}$ is the mean similarity across all local tokens. The final training objective integrates hierarchical alignment and hard-negative isolation:

\begin{equation}
L_{\text{total}} = \lambda_{gs} L_{gs} + \lambda_{h} \sum_{l \in P} \left( L_{ls}^{(l)} + L_{hi}^{(l)} \right) + L_{vpr}.
\end{equation}

\subsection{Intelligent DQ-pilot}

The DQ-pilot acts as a strategic visual reasoner, trained to formulate discriminative questions that enhance retrieval performance. Its training proceeds in two progressive stages: (1) Supervised Fine-Tuning (SFT) to establish foundational reasoning abilities, and (2) Reinforcement Learning (GRPO) to refine its question-generation strategy with task-aligned rewards.

\noindent\textbf{Supervised Fine-Tuning (SFT).}
In the first training stage, DQ-pilot is fine-tuned on a carefully selected DQ-cities-20k subset of the DQ-Cities dataset using the standard next-token prediction objective. Each training instance corresponds to a single dialogue turn, where the input consists of the current dialogue history \(Q_i\), the associated candidate set represented by \texttt{<image>} tokens, and an instruction that specifies the Questioner’s reasoning goal and response format. The output is a structured reasoning trace followed by a well-formed discriminative question that effectively differentiates visually similar locations. Through this next-token prediction process, DQ-pilot learns to connect accumulated dialogue context with spatial ambiguity and to formulate questions that progressively guide the retriever toward the correct place. This stage establishes the model’s foundational reasoning and dialogue abilities, providing a solid initialization for subsequent reinforcement refinement.

\noindent\textbf{Reinforcement Learning via GRPO.}
To further enhance strategic behavior beyond imitation, we refine the SFT-initialized model on the more challenging DQ-cities-10k subset using GRPO reinforcement learning.

\begin{itemize}[leftmargin=1.5em]
    \item \noindent\textbf{Format Reward ($R_{\text{fmt}}$).} To ensure consistent reasoning structure and interpretability, we define a binary reward verifying adherence to the required \texttt{<think></think><question></question>} template:
    \begin{equation}
    R_{\text{fmt}}(y) =
    \begin{cases}
    1, & \text{if } y \text{ matches the required format},\\
    0, & \text{otherwise.}
    \end{cases}
    \end{equation}
    
    \item \textbf{Retrieval Reward (\(R_{\text{prg}}\)).}  
    We reuse the Positional Retrieval Gain (PRG) from Sec~\ref{sec:ddi_sampling} as a task-aligned measure of how effectively a generated question improves localization.  
    Given the retriever’s updated ranks at round \(t\), the retrieval reward is defined as
    \begin{equation}
    R_{\text{prg}} = \mathrm{PRG}_t,
    \end{equation}
    which directly quantifies retrieval improvement induced by the model’s question.
\end{itemize}

\textbf{Final Objective.}  
The scalar reward used for GRPO optimization is a weighted combination of these two components:
\begin{equation}
R = \alpha\,R_{\text{prg}} + \beta\,R_{\text{fmt}},
\end{equation}
where \(\alpha, \beta > 0\) balance task performance and structural consistency.  
This reinforcement phase encourages DQ-Pilot to move beyond supervised imitation—learning to generate concise, discriminative, and retrieval-effective questions that actively steer the reasoning process within DlgPR.

\begin{table*}[t]
    \centering
    \caption{Static retrieval performance using fine-grained long descriptions across five representative regions. The best metrics are shown in \textbf{\textcolor{red}{red bold}}.}
    \label{tab:city_results}
    \renewcommand{\arraystretch}{1}
    \setlength{\tabcolsep}{3pt}
    \begin{tabularx}{\textwidth}{@{}l|*{3}{>{\centering\arraybackslash}X}||*{3}{>{\centering\arraybackslash}X}||*{3}{>{\centering\arraybackslash}X}||*{3}{>{\centering\arraybackslash}X}||*{3}{>{\centering\arraybackslash}X}}
    \toprule
    \multirow{2}{*}{Method}
      & \multicolumn{3}{c||}{\textbf{LosAngeles}}
      & \multicolumn{3}{c||}{\textbf{BuenosAires}}
      & \multicolumn{3}{c||}{\textbf{MexicoCity}}
      & \multicolumn{3}{c||}{\textbf{Osaka}}
      & \multicolumn{3}{c}{\textbf{PRG}} \\
    \cline{2-16}
      & \small{R@1} & \small{R@5} & \small{R@10}
      & \small{R@1} & \small{R@5} & \small{R@10}
      & \small{R@1} & \small{R@5} & \small{R@10}
      & \small{R@1} & \small{R@5} & \small{R@10}
      & \small{R@1} & \small{R@5} & \small{R@10} \\
    \hline
    
CLIP~\cite{radford2021learning} & 41.0 & 60.2 & 68.2 & 49.3 & 70.0 & 75.9 & 52.2 & 69.3 & 74.2 & 54.0 & 71.8 & 76.7 & 64.0 & 77.5 & 83.7 \\
Long-CLIP~\cite{zhang2024long} & 46.8 & 65.9 & 73.1 & 54.9 & 75.7 & 80.1 & 57.0 & 74.9 & 79.0 & 59.2 & 76.5 & 82.0 & 69.1 & 82.6 & 88.1 \\
FG-CLIP~\cite{xie2025fg} & 56.1 & 76.6 & 83.2 & 65.2 & 84.0 & 89.1 & 66.9 & 83.5 & 88.0 & 68.8 & 85.3 & 90.5 & 78.2 & 91.0 & 96.6 \\
Flair~\cite{xiao2025flair} & 57.9 & 76.2 & 79.9 & 66.1 & 85.2 & 90.3 & 66.5 & 83.2 & 88.2 & 68.3 & 85.1 & 89.2 & 78.1 & 89.1 & 95.9 \\
CMPL(Ours) & \textbf{\textcolor{red}{71.9}} & \textbf{\textcolor{red}{88.3}} & \textbf{\textcolor{red}{92.9}} & \textbf{\textcolor{red}{69.3}} & \textbf{\textcolor{red}{89.3}} & \textbf{\textcolor{red}{93.4}} & \textbf{\textcolor{red}{69.5}} & \textbf{\textcolor{red}{87.3}} & \textbf{\textcolor{red}{92.4}} & \textbf{\textcolor{red}{72.5}} & \textbf{\textcolor{red}{90.2}} & \textbf{\textcolor{red}{94.8}} & \textbf{\textcolor{red}{82.5}} & \textbf{\textcolor{red}{95.1}} & \textbf{\textcolor{red}{97.4}} \\
    \bottomrule
    \end{tabularx}
    \vspace{-0.3cm}
\end{table*}
\begin{table}[h]
\centering
\caption{Ablation study on CMPL retriever components. The average value of per-city tests across five representative regions. The best metrics are shown in \textbf{\textcolor{red}{red bold}}.}
\label{tab:retriever-ablation}
\begin{tabular}{lccc}
\toprule
\textbf{Configurations} & \textbf{R@1} & \textbf{R@5} & \textbf{R@10} \\
\midrule
Baseline & 71.6 & 88.0 & 95.3 \\
+ Token Selection & 71.9 & 88.5 & 95.9 \\
+ Progressive hsdm & 72.3 & 89.0 & 96.4 \\
+ HI (Hard-negative Isolation) & 72.7 & 89.5 & 97.0 \\
\midrule
\textbf{Full (All components)} & \textbf{\textcolor{red}{73.2}} & \textbf{\textcolor{red}{90.0}} & \textbf{\textcolor{red}{97.5}} \\
\bottomrule
\vspace{-0.5cm}
\end{tabular}
\end{table}

\begin{table}[h]
\centering
\caption{Ablation studies on key components of our DQ-pilot’s learning strategy. Final 5-round results are reported. The best metrics are shown in \textbf{\textcolor{red}{red bold}}.}
\label{tab:ablation}
\begin{tabular}{lcc}
\toprule
Setting                     & R@1 & R@5  \\ 
\midrule
DQ-pilot                 & \textbf{\textcolor{red}{60.5}}   & \textbf{\textcolor{red}{77.8}}     \\
w/o DDI Curriculum (Random)     & 59.6 & 77.2   \\
w/o GRPO (SFT-30k)         & 59.1 & 76.6     \\
w/o GRPO (SFT)         & 58.1   & 75.9   \\
\bottomrule
\vspace{-0.8cm}
\end{tabular}
\end{table}

\section{Experiments}
\label{sec:experiments}

\subsection{Experimental Setup}

\textbf{Dataset.}
All the experiments are conducted on our proposed DQ-cities dataset, and the evaluation is carried out for five representative cities from various continents. Each sample begins with a vague initial description (in the 0th round), and the questioner completes the retrieval through iterative dialogues. Table~\ref{tab:city_results_modified} reports results up to Round 5. 
\textbf{Evaluation Metrics.}
We evaluate the performance using cumulative Recall@K up to round $r$, where $k \in \{1, 5\}$, as the primary evaluation metric. In addition, the BRI index ~\cite{lee2024interactive} is introduced as an indicator to measure the efficiency of each round of questioning.
\noindent\textbf{Implementation Details.}
Our CMPL adopts a CLIP ViT-B/16 backbone and is trained using the proposed CMPL framework with fine-grained long descriptions as input, the number of learnable queries for each layer is set to 16. To handle long texts, we apply linear interpolation to the positional embeddings of tokens that exceed the original context length in the text encoder. The DQ-pilot is based on Qwen2.5-VL-7B-Instruct, fine-tuned with LoRA for parameter-efficient adaptation training follows our curriculum learning strategy : (1) Supervised Fine-Tuning (SFT) on low-DDI samples, and (2) GRPO-based reinforcement optimization on high-DDI samples. All the experiments are conducted on two A100s.

\subsection{Main Results}

\textbf{Interactive Reasoning Retrieval.}
As summarized in Table~\ref{tab:city_results_modified}, our fine-tuned DQ-pilot markedly surpasses both the original Qwen2.5-VL series and prior interactive retrieval methods. Compared to its 7B backbone, our model improves R@1 by 9.2\% and 13.4\% after 3 and 5 dialogue rounds, respectively, and even outperforms the much larger Qwen2.5-VL-72B by 7.3\%. This indicates that our progressive alignment strategy and reward-optimized training effectively boost interactive reasoning beyond mere model scaling. We further include specialized interactive retrievers, PlugIR~\cite{lee2024interactive}  as reference baselines to compare against established dialogue-driven retrieval pipelines. Our method achieves substantial gains in both R@1 and R@5 while maintaining the lowest BRI score, demonstrating superior interaction efficiency. It's also worth noting that the results highlight the advantage of our \textbf{SFT+GRPO} fine-tuning strategy: SFT provides structured reasoning alignment from supervised dialogues, while GRPO further promotes the model to achieve deeper reasoning. For more examples, please refer to the appendix.

\noindent\textbf{Retriever Performance.}
We also evaluate the retriever under ideal conditions using complete long descriptions, representing the upper bound of static retrieval. As shown in Table~\ref{tab:city_results}, our retriever significantly surpasses the Clip-based models including the recent state-of-the-art fine-grained image-text alignment models, validating its strong fine-grained cross-modal alignment.

\subsection{Ablation Studies}



To quantify the contribution of each core component, we conducted detailed ablation experiments. For DQ-pilot, regarding the Discriminative Difficulty-Aware Sampling strategy, we conducted two additional sets of control experiments: The first group trained the SFT model using the combined dataset of the two parts, while the second group performed SFT + GRPO training with the same sample quantity using a random sampling strategy. The results in Table \ref{tab:ablation} demonstrated that compared to imitation learning (SFT), the GRPO strategy could guide the model to perform deeper reasoning, and the curriculum setting guided by DDI was reasonable.
For CMPL, Table~\ref{tab:retriever-ablation} reports the results of the ablation experiments. Here, Baseline  indicates the use of only the $L_{gs}$ loss. We sequentially add the vpr loss and salient feature selection to the baseline to illustrate the importance of the salient location patches in the scene localization task. Then, we apply the multi-layer progressive CMPL, local-sdm loss, and HI loss, achieving the performance, which indicates that fully exploiting the alignment between fine-grained features is necessary.




\section{Conclusion}

We present DlgPR, a new paradigm that transforms traditional static geo-localization into an interactive, reasoning-driven process. Built upon our large-scale benchmark DQ-Cities and a curriculum guided by the DDI, our reasoning framework featuring the intelligent questioner DQ-Pilot—learns to iteratively refine spatial understanding through dialogue. Extensive experiments demonstrate that this interactive reasoning approach significantly enhances localization robustness and efficiency, highlighting the importance of active questioning for real-world geo-localization. In future work, we plan to explore more adaptive dialogue policies, tighter retriever–questioner co-training, and real-time deployment strategies for embodied agents in open environments.


{
    \small
    \bibliographystyle{ieeenat_fullname}
    \bibliography{main}

@String(CVPR= {IEEE Conf. Comput. Vis. Pattern Recog.})

@String(ECCV= {Eur. Conf. Comput. Vis.})

@String(AAAI = {AAAI})

@String(CVPR  = {CVPR})

@String(ECCV  = {ECCV})

@inproceedings{tian2024loc4plan,
  title={Loc4plan: Locating before planning for outdoor vision and language navigation},
  author={Tian, Huilin and Meng, Jingke and Zheng, Wei-Shi and Li, Yuan-Ming and Yan, Junkai and Zhang, Yunong},
  booktitle={Proceedings of the 32nd ACM International Conference on Multimedia},
  pages={4073--4081},
  year={2024}
}

@inproceedings{sarlin2019coarse,
  title={From coarse to fine: Robust hierarchical localization at large scale},
  author={Sarlin, Paul-Edouard and Cadena, Cesar and Siegwart, Roland and Dymczyk, Marcin},
  booktitle={Proceedings of the IEEE/CVF conference on computer vision and pattern recognition},
  pages={12716--12725},
  year={2019}
}

@article{wang2023fine,
  title={Fine-grained cross-view geo-localization using a correlation-aware homography estimator},
  author={Wang, Xiaolong and Xu, Runsen and Cui, Zhuofan and Wan, Zeyu and Zhang, Yu},
  journal={Advances in Neural Information Processing Systems},
  volume={36},
  pages={5301--5319},
  year={2023}
}

@article{ye2024skydiffusion,
  title={Skydiffusion: Street-to-satellite image synthesis with diffusion models and bev paradigm},
  author={Ye, Junyan and He, Jun and Li, Weijia and Lv, Zhutao and Yu, Jinhua and Yang, Haote and He, Conghui},
  journal={arXiv e-prints},
  pages={arXiv--2408},
  year={2024}
}

@inproceedings{netvlad,
  title={NetVLAD: CNN architecture for weakly supervised place recognition},
  author={Arandjelovic, Relja and Gronat, Petr and Torii, Akihiko and Pajdla, Tomas and Sivic, Josef},
  booktitle={Proceedings of the IEEE conference on computer vision and pattern recognition},
  pages={5297--5307},
  year={2016}
}

@InProceedings{salad,
    author    = {Izquierdo, Sergio and Civera, Javier},
    title     = {Optimal Transport Aggregation for Visual Place Recognition},
    booktitle = {Proceedings of the IEEE/CVF Conference on Computer Vision and Pattern Recognition (CVPR)},
    month     = {June},
    year      = {2024},
}

@inproceedings{wang2019multi,
  title={Multi-similarity loss with general pair weighting for deep metric learning},
  author={Wang, Xun and Han, Xintong and Huang, Weilin and Dong, Dengke and Scott, Matthew R},
  booktitle={Proceedings of the IEEE/CVF conference on computer vision and pattern recognition},
  pages={5022--5030},
  year={2019}
}

@inproceedings{wang2013theoretical,
  title={A theoretical analysis of NDCG type ranking measures},
  author={Wang, Yining and Wang, Liwei and Li, Yuanzhi and He, Di and Liu, Tie-Yan},
  booktitle={Conference on learning theory},
  pages={25--54},
  year={2013},
  organization={PMLR}
}

@article{xie2025fg,
  title={FG-CLIP: Fine-Grained Visual and Textual Alignment},
  author={Xie, Chunyu and Wang, Bin and Kong, Fanjing and Li, Jincheng and Liang, Dawei and Zhang, Gengshen and Leng, Dawei and Yin, Yuhui},
  journal={arXiv preprint arXiv:2505.05071},
  year={2025}
}

@inproceedings{xiao2025flair,
  title={Flair: Vlm with fine-grained language-informed image representations},
  author={Xiao, Rui and Kim, Sanghwan and Georgescu, Mariana-Iuliana and Akata, Zeynep and Alaniz, Stephan},
  booktitle={Proceedings of the Computer Vision and Pattern Recognition Conference},
  pages={24884--24894},
  year={2025}
}

@inproceedings{zhang2024long,
  title={Long-clip: Unlocking the long-text capability of clip},
  author={Zhang, Beichen and Zhang, Pan and Dong, Xiaoyi and Zang, Yuhang and Wang, Jiaqi},
  booktitle={European conference on computer vision},
  pages={310--325},
  year={2024},
  organization={Springer}
}

@inproceedings{radford2021learning,
  title={Learning transferable visual models from natural language supervision},
  author={Radford, Alec and Kim, Jong Wook and Hallacy, Chris and Ramesh, Aditya and Goh, Gabriel and Agarwal, Sandhini and Sastry, Girish and Askell, Amanda and Mishkin, Pamela and Clark, Jack and others},
  booktitle={International conference on machine learning},
  pages={8748--8763},
  year={2021},
  organization={PmLR}
}

@InProceedings{boq,
    author    = {Ali-bey, Amar and Chaib-draa, Brahim and Gigu\`ere, Philippe},
    title     = {{BoQ}: A Place is Worth a Bag of Learnable Queries},
    booktitle = {Proceedings of the IEEE/CVF Conference on Computer Vision and Pattern Recognition},
    year      = {2024},
    pages     = {17794-17803}
}

@article{supervlad,
  title={SuperVLAD: Compact and robust image descriptors for visual place recognition},
  author={Lu, Feng and Zhang, Xinyao and Ye, Canming and Dong, Shuting and Zhang, Lijun and Lan, Xiangyuan and Yuan, Chun},
  journal={Advances in Neural Information Processing Systems},
  volume={37},
  pages={5789--5816},
  year={2024}
}

@article{emvp,
  title={EMVP: Embracing Visual Foundation Model for Visual Place Recognition with Centroid-Free Probing},
  author={Qiu, Qibo and Zhang, Shun and Gao, Haiming and Yang, Honghui and Ying, Haochao and Wang, Wenxiao and He, Xiaofei},
  journal={Advances in Neural Information Processing Systems},
  volume={37},
  pages={120928--120950},
  year={2024}
}

@inproceedings{selavpr,
  title={Towards Seamless Adaptation of Pre-trained Models for Visual Place Recognition},
  author={Lu, Feng and Zhang, Lijun and Lan, Xiangyuan and Dong, Shuting and Wang, Yaowei and Yuan, Chun},
  booktitle={The Twelfth International Conference on Learning Representations},
  year={2024}
}

@inproceedings{FoL,
  title={Focus on Local: Finding Reliable Discriminative Regions for Visual Place Recognition},
  author={Wang, Changwei and Chen, Shunpeng and Song, Yukun and Xu, Rongtao and Zhang, Zherui and Zhang, Jiguang and Yang, Haoran and Zhang, Yu and Fu, Kexue and Du, Shide and others},
  booktitle={Proceedings of the AAAI Conference on Artificial Intelligence},
  volume={39},
  number={7},
  pages={7536--7544},
  year={2025}
}

@inproceedings{ye2025cross,
  title={Where am i? cross-view geo-localization with natural language descriptions},
  author={Ye, Junyan and Lin, Honglin and Ou, Leyan and Chen, Dairong and Wang, Zihao and Zhu, Qi and He, Conghui and Li, Weijia},
  booktitle={Proceedings of the IEEE/CVF International Conference on Computer Vision},
  pages={5890--5900},
  year={2025}
}

@inproceedings{kolmet2022text2pos,
  title={Text2pos: Text-to-point-cloud cross-modal localization},
  author={Kolmet, Manuel and Zhou, Qunjie and O{\v{s}}ep, Aljo{\v{s}}a and Leal-Taix{\'e}, Laura},
  booktitle={Proceedings of the IEEE/CVF Conference on Computer Vision and Pattern Recognition},
  pages={6687--6696},
  year={2022}
}

@article{tao2025textinplace,
  title={TextInPlace: Indoor Visual Place Recognition in Repetitive Structures with Scene Text Spotting and Verification},
  author={Tao, Huaqi and Liu, Bingxi and Chen, Calvin and Huang, Tingjun and Li, He and Cui, Jinqiang and Zhang, Hong},
  journal={arXiv preprint arXiv:2503.06501},
  year={2025}
}

@inproceedings{chu2024towards,
  title={Towards natural language-guided drones: GeoText-1652 benchmark with spatial relation matching},
  author={Chu, Meng and Zheng, Zhedong and Ji, Wei and Wang, Tingyu and Chua, Tat-Seng},
  booktitle={European Conference on Computer Vision},
  pages={213--231},
  year={2024},
  organization={Springer}
}

@inproceedings{xia2024text2loc,
  title={Text2loc: 3d point cloud localization from natural language},
  author={Xia, Yan and Shi, Letian and Ding, Zifeng and Henriques, Joao F and Cremers, Daniel},
  booktitle={Proceedings of the IEEE/CVF conference on computer vision and pattern recognition},
  pages={14958--14967},
  year={2024}
}

@article{lyu2024tell,
  title={Tell me where you are: Multimodal llms meet place recognition},
  author={Lyu, Zonglin and Zhang, Juexiao and Lu, Mingxuan and Li, Yiming and Feng, Chen},
  journal={arXiv preprint arXiv:2406.17520},
  year={2024}
}

@article{wang2024lvlm,
  title={LVLM-empowered Multi-modal Representation Learning for Visual Place Recognition},
  author={Wang, Teng and Meng, Lingquan and Cheng, Lei and Sun, Changyin},
  journal={arXiv preprint arXiv:2407.06730},
  year={2024}
}

@inproceedings{hong2019textplace,
  title={TextPlace: Visual place recognition and topological localization through reading scene texts},
  author={Hong, Ziyang and Petillot, Yvan and Lane, David and Miao, Yishu and Wang, Sen},
  booktitle={Proceedings of the IEEE/CVF International Conference on Computer Vision},
  pages={2861--2870},
  year={2019}
}

@inproceedings{hu2024progeo,
  title={Progeo: Generating prompts through image-text contrastive learning for visual geo-localization},
  author={Hu, Jingqi and Mao, Chen and Tan, Chong and Li, Hui and Liu, Hong and Zheng, Min},
  booktitle={International Conference on Artificial Neural Networks},
  pages={448--462},
  year={2024},
  organization={Springer}
}

@inproceedings{madasu2022learning,
  title={Learning to retrieve videos by asking questions},
  author={Madasu, Avinash and Oliva, Junier and Bertasius, Gedas},
  booktitle={Proceedings of the 30th ACM International Conference on Multimedia},
  pages={356--365},
  year={2022}
}

@article{lee2024interactive,
  title={Interactive text-to-image retrieval with large language models: A plug-and-play approach},
  author={Lee, Saehyung and Yu, Sangwon and Park, Junsung and Yi, Jihun and Yoon, Sungroh},
  journal={arXiv preprint arXiv:2406.03411},
  year={2024}
}

@article{lu2025llava,
  title={LLaVA-ReID: Selective Multi-image Questioner for Interactive Person Re-Identification},
  author={Lu, Yiding and Yang, Mouxing and Peng, Dezhong and Hu, Peng and Lin, Yijie and Peng, Xi},
  journal={arXiv preprint arXiv:2504.10174},
  year={2025}
}

@inproceedings{garg2022seqmatchnet,
  title={Seqmatchnet: Contrastive learning with sequence matching for place recognition \& relocalization},
  author={Garg, Sourav and Vankadari, Madhu and Milford, Michael},
  booktitle={Conference on Robot Learning},
  pages={429--443},
  year={2022},
  organization={PMLR}
}

@inproceedings{li2023omnicity,
  title={Omnicity: Omnipotent city understanding with multi-level and multi-view images},
  author={Li, Weijia and Lai, Yawen and Xu, Linning and Xiangli, Yuanbo and Yu, Jinhua and He, Conghui and Xia, Gui-Song and Lin, Dahua},
  booktitle={Proceedings of the IEEE/CVF Conference on Computer Vision and Pattern Recognition},
  pages={17397--17407},
  year={2023}
}

@inproceedings{shi2020looking,
  title={Where am i looking at? joint location and orientation estimation by cross-view matching},
  author={Shi, Yujiao and Yu, Xin and Campbell, Dylan and Li, Hongdong},
  booktitle={Proceedings of the IEEE/CVF Conference on Computer Vision and Pattern Recognition},
  pages={4064--4072},
  year={2020}
}

@inproceedings{xia2024adapting,
  title={Adapting fine-grained cross-view localization to areas without fine ground truth},
  author={Xia, Zimin and Shi, Yujiao and Li, Hongdong and FP Kooij, Julian},
  booktitle={European Conference on Computer Vision},
  pages={397--415},
  year={2024},
  organization={Springer}
}

@inproceedings{lee2021cosmo,
  title={Cosmo: Content-style modulation for image retrieval with text feedback},
  author={Lee, Seungmin and Kim, Dongwan and Han, Bohyung},
  booktitle={Proceedings of the IEEE/CVF Conference on Computer Vision and Pattern Recognition},
  pages={802--812},
  year={2021}
}

@inproceedings{cai2021ask,
  title={Ask\&confirm: active detail enriching for cross-modal retrieval with partial query},
  author={Cai, Guanyu and Zhang, Jun and Jiang, Xinyang and Gong, Yifei and He, Lianghua and Yu, Fufu and Peng, Pai and Guo, Xiaowei and Huang, Feiyue and Sun, Xing},
  booktitle={Proceedings of the IEEE/CVF International Conference on Computer Vision},
  pages={1835--1844},
  year={2021}
}

@article{gsv,
  title={Gsv-cities: Toward appropriate supervised visual place recognition},
  author={Ali-bey, Amar and Chaib-draa, Brahim and Gigu{\`e}re, Philippe},
  journal={Neurocomputing},
  volume={513},
  pages={194--203},
  year={2022},
  publisher={Elsevier}
}

@article{kovashka2015whittlesearch,
  title={Whittlesearch: Interactive image search with relative attribute feedback},
  author={Kovashka, Adriana and Parikh, Devi and Grauman, Kristen},
  journal={International Journal of Computer Vision},
  volume={115},
  number={2},
  pages={185--210},
  year={2015},
  publisher={Springer}
}

@inproceedings{liang2023simple,
  title={Simple baselines for interactive video retrieval with questions and answers},
  author={Liang, Kaiqu and Albanie, Samuel},
  booktitle={Proceedings of the IEEE/CVF International Conference on Computer Vision},
  pages={11091--11101},
  year={2023}
}

@inproceedings{zhu2024enhancing,
  title={Enhancing interactive image retrieval with query rewriting using large language models and vision language models},
  author={Zhu, Hongyi and Huang, Jia-Hong and Rudinac, Stevan and Kanoulas, Evangelos},
  booktitle={Proceedings of the 2024 International Conference on Multimedia Retrieval},
  pages={978--987},
  year={2024}
}

@inproceedings{jiang2023cross,
  title={Cross-modal implicit relation reasoning and aligning for text-to-image person retrieval},
  author={Jiang, Ding and Ye, Mang},
  booktitle={Proceedings of the IEEE/CVF conference on computer vision and pattern recognition},
  pages={2787--2797},
  year={2023}
}

@article{levy2023chatting,
  title={Chatting makes perfect: Chat-based image retrieval},
  author={Levy, Matan and Ben-Ari, Rami and Darshan, Nir and Lischinski, Dani},
  journal={Advances in Neural Information Processing Systems},
  volume={36},
  pages={61437--61449},
  year={2023}
}

@article{guo2025deepseek,
  title={Deepseek-r1: Incentivizing reasoning capability in llms via reinforcement learning},
  author={Guo, Daya and Yang, Dejian and Zhang, Haowei and Song, Junxiao and Zhang, Ruoyu and Xu, Runxin and Zhu, Qihao and Ma, Shirong and Wang, Peiyi and Bi, Xiao and others},
  journal={arXiv preprint arXiv:2501.12948},
  year={2025}
}

@article{jaech2024openai,
  title={Openai o1 system card},
  author={Jaech, Aaron and Kalai, Adam and Lerer, Adam and Richardson, Adam and El-Kishky, Ahmed and Low, Aiden and Helyar, Alec and Madry, Aleksander and Beutel, Alex and Carney, Alex and others},
  journal={arXiv preprint arXiv:2412.16720},
  year={2024}
}

@article{shao2024deepseekmath,
  title={Deepseekmath: Pushing the limits of mathematical reasoning in open language models},
  author={Shao, Zhihong and Wang, Peiyi and Zhu, Qihao and Xu, Runxin and Song, Junxiao and Bi, Xiao and Zhang, Haowei and Zhang, Mingchuan and Li, YK and Wu, Yang and others},
  journal={arXiv preprint arXiv:2402.03300},
  year={2024}
}

@article{shen2025vlm,
  title={Vlm-r1: A stable and generalizable r1-style large vision-language model},
  author={Shen, Haozhan and Liu, Peng and Li, Jingcheng and Fang, Chunxin and Ma, Yibo and Liao, Jiajia and Shen, Qiaoli and Zhang, Zilun and Zhao, Kangjia and Zhang, Qianqian and others},
  journal={arXiv preprint arXiv:2504.07615},
  year={2025}
}

@article{liu2025visual,
  title={Visual-rft: Visual reinforcement fine-tuning},
  author={Liu, Ziyu and Sun, Zeyi and Zang, Yuhang and Dong, Xiaoyi and Cao, Yuhang and Duan, Haodong and Lin, Dahua and Wang, Jiaqi},
  journal={arXiv preprint arXiv:2503.01785},
  year={2025}
}

@inproceedings{ye2024cross,
  title={Cross-view image geo-localization with Panorama-BEV Co-Retrieval Network},
  author={Ye, Junyan and Lv, Zhutao and Li, Weijia and Yu, Jinhua and Yang, Haote and Zhong, Huaping and He, Conghui},
  booktitle={European Conference on Computer Vision},
  pages={74--90},
  year={2024},
  organization={Springer}
}

@article{ali2023global,
  title={Global proxy-based hard mining for visual place recognition},
  author={Ali-Bey, Amar and Chaib-draa, Brahim and Gigu{\`e}re, Philippe},
  journal={arXiv preprint arXiv:2302.14217},
  year={2023}
}

@InProceedings{salad-cm,
author="Izquierdo, Sergio and Civera, Javier",
title="Close, But Not There: Boosting Geographic Distance Sensitivity in Visual Place Recognition",
booktitle="Computer Vision -- ECCV 2024",
year="2025",
publisher="Springer Nature Switzerland",
address="Cham",
pages="240--257",
}

@inproceedings{deuser2023sample4geo,
  title={Sample4geo: Hard negative sampling for cross-view geo-localisation},
  author={Deuser, Fabian and Habel, Konrad and Oswald, Norbert},
  booktitle={Proceedings of the IEEE/CVF International Conference on Computer Vision},
  pages={16847--16856},
  year={2023}
}

@inproceedings{chen2024scene,
  title={“Where am I?” Scene Retrieval with Language},
  author={Chen, Jiaqi and Barath, Daniel and Armeni, Iro and Pollefeys, Marc and Blum, Hermann},
  booktitle={European Conference on Computer Vision},
  pages={201--220},
  year={2024},
  organization={Springer}
}

@article{shang2025bridging,
  title={Bridging text and vision: A multi-view text-vision registration approach for cross-modal place recognition},
  author={Shang, Tianyi and Li, Zhenyu and Xu, Pengjie and Qiao, Jinwei and Chen, Gang and Ruan, Zihan and Hu, Weijun},
  journal={arXiv preprint arXiv:2502.14195},
  year={2025}
}

@article{zhang2024navid,
  title={Navid: Video-based vlm plans the next step for vision-and-language navigation},
  author={Zhang, Jiazhao and Wang, Kunyu and Xu, Rongtao and Zhou, Gengze and Hong, Yicong and Fang, Xiaomeng and Wu, Qi and Zhang, Zhizheng and Wang, He},
  journal={arXiv preprint arXiv:2402.15852},
  year={2024}
}

@inproceedings{SAGE,
    title={{SAGE}: Spatial-visual Adaptive Graph Exploration for Efficient Visual Place Recognition},
    author={Shunpeng Chen and Changwei Wang and Rongtao Xu and Xingtian Pei and Yukun Song and Jinzhou Lin and Wenhao Xu and Jingyi Zhang and Li Guo and Shibiao Xu},
    booktitle={The Fourteenth International Conference on Learning Representations},
    year={2026},
    url={https://openreview.net/forum?id=DCpbEXqPvS}
}

@article{chen2026region,
  title={Region Matters: Efficient and Reliable Region-Aware Visual Place Recognition},
  author={Chen, Shunpeng and Song, Yukun and Wang, Changwei and Xu, Rongtao and Fu, Kexue and Gao, Longxiang and Guo, Li and Wang, Ruisheng and Xu, Shibiao},
  journal={arXiv preprint arXiv:2604.22390},
  year={2026}
}
}


\end{document}